\documentclass[10pt,twocolumn,letterpaper]{article}

\usepackage{cvpr}
\usepackage{times}
\usepackage{epsfig}
\usepackage{graphicx}
\usepackage{amsmath}
\usepackage{amssymb}

\usepackage[pagebackref=true,breaklinks=true,letterpaper=true,colorlinks,bookmarks=false]{hyperref}

\cvprfinalcopy 

\def\cvprPaperID{9} 

\ifcvprfinal\fi
\begin{document}

\title{Image-based Vehicle Re-identification Model with Adaptive  Attention Modules and Metadata Re-ranking}

\author{Quang Truong, Hy Dang, Zhankai Ye, Minh Nguyen, Bo Mei\\
Texas Christian University\\
Fort Worth, TX\\
{\tt\small \{quang.truong, hy.dang, zhankai.ye, minh.d.nguyen,  b.mei\}@tcu.edu}
}

\maketitle

\begin{abstract}
   Vehicle Re-identification is a challenging task due to intra-class variability and inter-class similarity across non-overlapping cameras. To tackle these problems, recently proposed methods require additional annotation to extract more features for false positive image exclusion. In this paper, we propose a model powered by adaptive attention modules that requires fewer label annotations but still outperforms the previous models. We also include a re-ranking method that takes account of the importance of metadata feature embeddings in our paper. The proposed method is evaluated on CVPR AI City Challenge 2020 dataset and achieves mAP of 37.25\% in Track 2.
\end{abstract}

\section{Introduction}

In recent years, computer vision has achieved accomplishments across its sub-fields thanks to the continuing development of Convolutional Neural Network (CNN). Among sub-fields of computer vision, object re-identification has gained attention lately due to several technical difficulties. The first challenge is intra-class variability. Because of illumination conditions, obstacles, and occlusions, an object may appear different across non-overlapping cameras. The second challenge is the inter-class similarity. Two objects may share similar looks, such as identical twins or cars from the same manufacturing process. Unlike image classification whose task is to classify images based on visual contents, object re-identification demands a robust system to respond to local features and global features. Local features involve differentiating two objects with similar viewpoints. In contrast, global features involve clustering images that belong to the same objects, regardless of viewpoints. Re-identification systems also have to possess a good generalization ability to deal with unseen features due to plenty of object variations.

Initially, most of the research projects about re-identification focus on person re-identification, and vehicle re-identification has adopted the previous contributions successfully despite the difference of domains \cite{hermans2017defense, kumar2019vehicle, luo2019bag, tang2019cityflow, suprem2020looking, liu2016large-scale, liu2016deep, chen2019deep, nguyen2019vehicle, wang2017orientation, Sankaranarayanan_2016, Huang_2019_CVPR_Workshops, Khorramshahi_2019_CVPR_Workshops}. However, the majority of these projects adopt the pre-trained ImageNet classification-specific models and perform transfer learning for the vehicle re-identification task. Our proposed method focuses on GLAMOR, a model designed for re-identification proposed by Suprem \etal \cite{suprem2020looking}, which proves that training from scratch with a smaller dataset ($36,935$ real images and $192,150$ synthetic images versus $14$M images of ImageNet) does not necessarily result in poorer performance. In fact, GLAMOR outperforms ResNet50 baseline with $7.9$\% mAP improvement \cite{tang2019cityflow}. We also propose a slight modification to $k$-reciprocal encoding re-ranking \cite{zhong2017reranking} so that it includes the metadata attributes during the re-ranking process. The remainder of the paper is structured as follows: Section $2$ reviews the related work, Section $3$ illustrates our proposed approach, Section $4$ focuses on our experiment, and Section $5$ draws a conclusion and discusses potential rooms for improvement to study the re-identification problem.
\section{Related Work}

Re-identification problems have been a challenging task in computer vision. Unlike image classification, where images are required to be classified into classes, re-identification is to identify a probe image in an image gallery. While image classification achieves successful results \cite{szegedy2015rethinking, wang2017residual, woo2018cbam} thanks to large popular datasets such as COCO \cite{lin2014microsoft} or ImageNet \cite{deng2009a}\cite{krizhevsky2017imagenet}, re-identification is yet to have sufficiently large datasets to train model. DukeMTMC \cite{Gou_2017_CVPR_Workshops} and Market-1501 \cite{zheng2015scalable} are datasets specifically for person re-identification, while Veri-776 \cite{liu2016large-scale} and VehicleID \cite{liu2016deep} are for vehicle re-identification. These datasets share a common disadvantage, which is the lack of images per identity. Intra-class variability and inter-class similarity are also common problems in re-identification due to diverse backgrounds or similar looks.

Novel approaches to overcome the above disadvantages have been proposed recently. Hermans \etal prove that triplet loss \cite{weinberger2009distance, schroff2015facenet, chen2019deep} is suitable for re-identification task since it optimizes the embedding space so that images with the same identity are closer to each other compared to those with different identities \cite{hermans2017defense}. Hermans \etal also propose the Batch Hard technique to select the hardest negative samples within a batch, minimizing the intra-class variability of an identity \cite{hermans2017defense, chen2019deep, kumar2019vehicle}.

Besides data mining techniques and alternative loss functions, there have been several efforts to implement new models designed for re-identification \cite{suprem2020looking, liu2016deep, hermans2017defense, wang2017orientation, Khorramshahi_2019_CVPR_Workshops}. Specifically, Suprem \etal focus on using attention-based regularizers \cite{suprem2020looking, Khorramshahi_2019_CVPR_Workshops} to extract more global and local features and ensure low sparsity of activations. Wang \etal utilize 20 key point locations to extract local features based on orientation thanks to attention mechanism, and then fuse the extracted features with global features for orientation-invariant feature embedding \cite{wang2017orientation}.

Re-ranking is also an important post-processing method that is worth considering in re-identification. Zhong \etal propose a re-ranking method that encodes the $k$-reciprocal nearest neighbors of a probe image to calculate $k$-reciprocal feature \cite{zhong2017reranking}. The Jaccard distance is then re-calculated and combined with the original distance to get the final distance. Khorramshahi \etal utilize triplet probabilistic embedding \cite{Sankaranarayanan_2016} proposed by Sankaranarayanan \etal to create similarity score for re-ranking task \cite{Khorramshahi_2019_CVPR_Workshops}. Huang \etal propose metadata distance, which uses classification confidence and confusion distance. Metadata distance is then combined with the original distance to get the final distance \cite{Huang_2019_CVPR_Workshops}.

\section{Proposed Approach}
\subsection{System Overview}
\begin{figure*}
\begin{center}
\includegraphics[width=\linewidth]{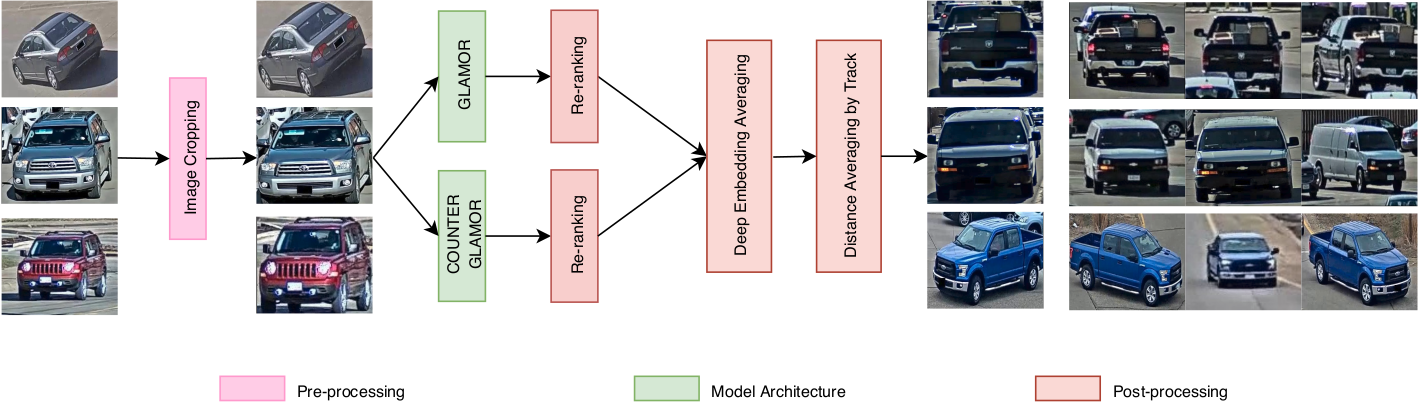}
\end{center}
   \caption{System Overview.}
\label{fig:system}
\end{figure*}

The overview of our system can be generalized into three main stages: pre-processing, deep embedding computing, and post-processing. The system is described in Figure~\ref{fig:system}.
Pre-processing is necessary since the bounding boxes of the provided dataset are loosely cropped. The loosely cropped images contain unnecessary information, which hinders the performance of our model.

The deep metric embedding module is a combination of two models, GLAMOR \cite{suprem2020looking} and Counter GLAMOR, that are trained on the provided dataset. The output of the module is a $W \times H$ distance matrix where $W$ represents the images in query and $H$ represents the images in the gallery.  Additional classifiers are also trained on the provided dataset to extract metadata attributes for further post-processing.

\begin{figure}[t]
\begin{center}
   \includegraphics[width=\linewidth]{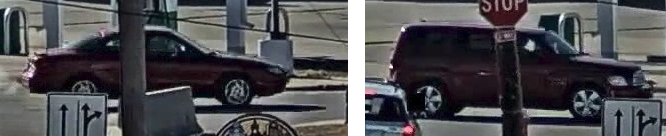}
\end{center}
   \caption{An example of two images with close embedding distance due to similar brightness, pose, color, and occlusion.}
\label{fig:close_embedding}
\end{figure}

Post-processing is essential in re-identification since it removes false-positive images at the top. Illumination conditions, vehicle poses, and other various factors affect the outputs negatively. Figure~\ref{fig:close_embedding} shows an example of two images with close embedding distance due to similarities in brightness, pose, color, and occlusion. 

\subsection{Pre-processing}
\subsubsection{Detectron2}

We adopt pretrained Detectron2 \cite{wu2019detectron2} on MS COCO dataset \cite{lin2014microsoft} to detect vehicle in an image and then to crop the bounding box out of the image. Detectron2 is a Facebook platform for object detection and segmentation that implements state-of-the-art object detection algorithms, including Mask R-CNN\cite{he2017maskrcnn}. We perform image cropping on training, query, and test sets, and then use the cropped images for training as well as evaluating models.

\subsubsection{Image Labeling for Vehicle Attribute Extractor}
As shown in Figure~\ref{fig:close_embedding}, car type does not match. Even though they have close embedding distance, the embedding distance is mostly affected by noise features. Therefore, vehicle metadata attributes should be extracted to eliminate undesired features such as obstacles in the background.\\ \\
We adopt pre-trained ResNeXt101\cite{xie2016aggregated, cadene2019} on ImageNet \cite{krizhevsky2017imagenet} for rapid convergence. We train ResNext101 to classify color and type.\\ \\
The given color labels and type labels do not reflect the training set. For example, the training set does not contain any orange cars. The number of cars per category is also unevenly distributed; there is a lack of RV or bus images. Therefore, we cluster types based on their common visual attributes. For types, we suggest having $6$ categories: small vehicle (sedan, hatchback, estate, and sports car), big vehicle (SUV and MPV), van, pickup, truck, and long car (bus and RV). For color, we exclude orange, pink, and purple.\\ \\
The query set and test set, however, do contain the excluded categories. Moreover, there are different cameras in the query and test sets (the training set is collected from $36$ cameras while the query and test sets are collected from $23$ cameras), so performing prediction on the query and test sets will eventually result in incorrect classification. Therefore, we extract the features before the last fully-connected layer and calculate the Euclidean distance between the query set and the test set for the re-ranking process.
\subsection{Deep Embedding Computing}

\begin{figure}[t]
\begin{center}
   \includegraphics[width=\linewidth]{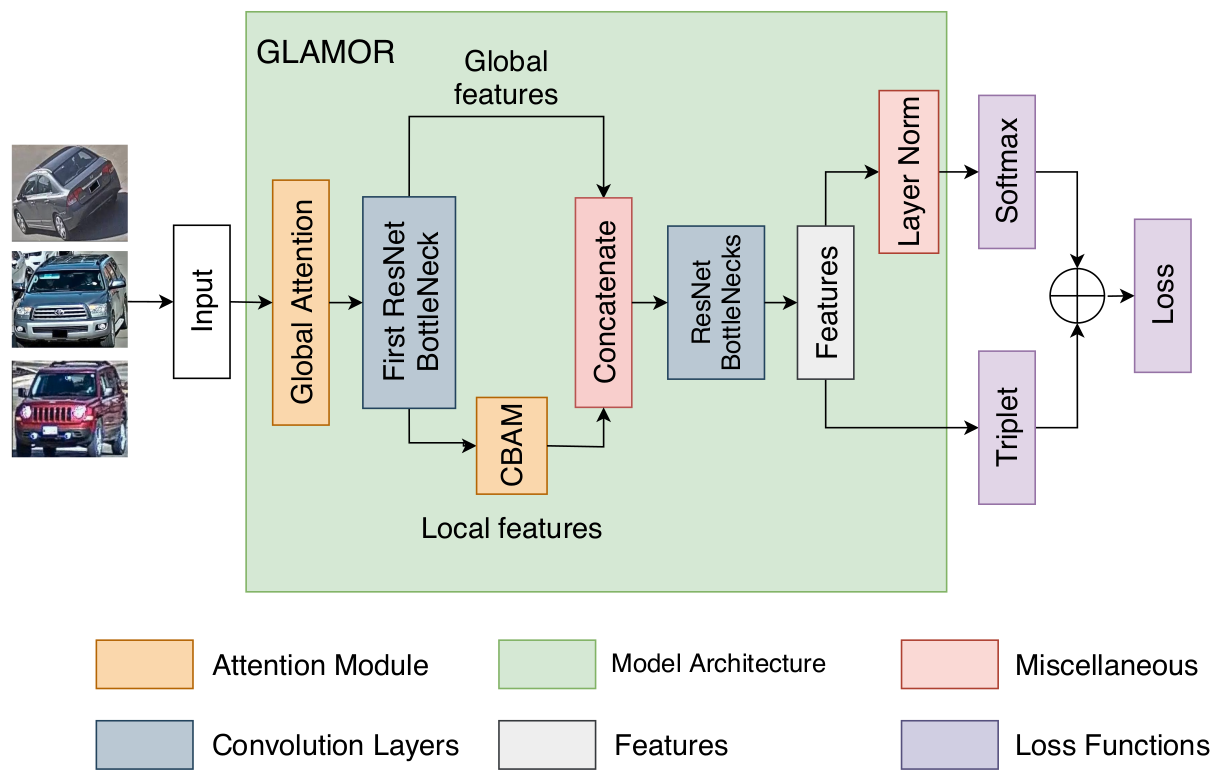}
\end{center}
   \caption{Architecture of GLAMOR.}
\label{fig:glamor}
\end{figure}

We adopt GLAMOR, an end-to-end ResNet50-backboned re-identification model powered by attention mechanism, proposed by Suprem \etal \cite{suprem2020looking}. GLAMOR introduces two modules. Global Attention Module reduces sparsity and enhances feature extraction. In the meantime, Local Attention Module extracts unsupervised part-based features. Unlike the original model, we have modified the model slightly to increase the performance. Instead of using the original Local Attention Module, we use Convolutional Block Attention Module (CBAM) as our local feature extractor because CBAM focuses on two principal dimensions: spatial and channel \cite{woo2018cbam, teng2019}. As a feature adaptive refinement module, CBAM learns effectively where to emphasize or suppress the information to be passed forward to the later convolutional blocks. The detailed architecture of GLAMOR is represented in Figure~\ref{fig:glamor}.

We also realize the loss of information in the current GLAMOR implementation at the concatenation step. Suprem \etal apply a channel-wise mask to combine global features and local features \cite{suprem2020looking}. However, only half of each is fed forward to later convolutional blocks. The sum of global features $F_G$ and local features $F_L$, where $F_G, F_L \in \mathbb{R}^{H \times W \times C}$, is calculated as follow:
\begin{equation} \label{eq:concat1}
    F = M_G \odot F_G + M_L \odot F_L,
\end{equation}
where $M_G, M_L \in \mathbb{R}^C$, $M_L = \Bar{M_G}$, and for each $m_i \in M_G$, $m_i = 0$ $ \forall i < \lfloor \frac{C}{2} \rfloor$ and $m_i = 1$ $\forall i \geq \lfloor \frac{C}{2} \rfloor$. Therefore, we propose another concatenation formula to counter the loss of information in Equation~(\ref{eq:concat1}) just by swapping the mask position:
\begin{equation} \label{eq:concat2}
    F = M_L \odot F_G + M_G \odot F_L.
\end{equation}

The concatenation formula in Equation~(\ref{eq:concat2}) is used for another GLAMOR. The distance embedding matrix of two GLAMORs is then averaged for the final result. The proposed method significantly increases the accuracy due to generalization and balancing effects.

The two models are trained separately on both synthetic data and training data. Training models on synthetic dataset helps models converge faster than training on the real dataset alone. Our models converge in $20-30$ epochs, while a pre-trained ResNet50 baseline model converges after $60$ epochs \cite{tang2019cityflow}. 

Our metric learning method is a combination of batch hard triplet loss \cite{weinberger2009distance, schroff2015facenet} and softmax loss with label smoothing \cite{szegedy2015rethinking}. The reason is that triplet loss is used for learning embeddings whereas softmax loss inteprets probability distributions of a list of potential outcomes. The combination loss is
\begin{equation}
    \mathcal{L}_{\text{TriSoft}} = \lambda_{\text{Triplet}} \cdot \mathcal{L}_{\text{Triplet}} + \lambda_{\text{Softmax}} \cdot \mathcal{L}_{\text{Softmax}}, 
\end{equation}{}
where $\lambda_{\text{Triplet}}$ and $\lambda_{\text{Softmax}}$ are hyperparameters that can be fine-tuned. The revised triplet loss proposed by FaceNet \cite{schroff2015facenet} is
\begin{equation}\label{eq:triplet}
    \mathcal{L}_{\text{Triplet}} = \sum_{\substack{a,p,n \\ y_a = y_p \neq y_n}} [m + D_{a,p} - D_{a,n}]_+ ,
\end{equation}
where $a,p,n$ are anchor, positive, and negative samples of a triplet, $D_{a,p}$ and $D_{a,n}$ are the distance from an anchor sample to a positive sample and to a negative sample, and $m$ is the margin constraint. The softmax with label smoothing proposed by Szegedy \etal \cite{szegedy2015rethinking} is
\begin{equation}\label{softmax}
\mathcal{L}_{\text{Softmax}} = \sum_{i=1}^N -q_i\log(p_i)
    \begin{cases}{}
  q_i = 0, y\neq i \nonumber\\
  q_i = 1,y = i \nonumber
\end{cases}
\end{equation}
and:
\begin{equation}
q_i = 
    \begin{cases}{}
  1 - \frac{N-1}{N}\varepsilon  & if \quad i = y\nonumber\\
  \varepsilon/N& \text{otherwise,} \nonumber
\end{cases}
\end{equation}
where $y$ is the ground truth ID label, $p_i$ is the ID prediction logits of class $i$, $N$ is the number of IDs in the dataset, and $\varepsilon$ is a hyperparameter to reduce over-confidence of classifiers.

\subsection{Post-processing}
\subsubsection{Re-ranking}
We adopt the re-ranking with $k$-reciprocal encoding method \cite{zhong2017reranking} proposed by Zhong \etal and modify the formula to include Euclidean distance embedding of metadata attributes. Given a probe image $p$ and a gallery image $g_i \in \mathcal{G}$ where $\mathcal{G}$ is gallery set, the revised original distance matrix is
\begin{equation}
    d^{\prime}(p, g_i) = d(p, g_i) + \sum \gamma_j \cdot D_j(p, g_i),
\end{equation}
where $d(p, g_i)$ is the original distance between $p$ and $g_i$, $\gamma_j$ is the hyperparameter of feature $j$ for fine-tuning, and $D_j(p, g_i)$ is the metadata distance between $p$ and $g_i$ of feature $j$. We then generate the $k$-reciprocal nearest neighbor set $\mathcal{R}$ and re-calculate the pairwise distance between the probe image $p$ and the gallery image $g_i$ using Jaccard distance and a more robust $k$-reciprocal nearest neighbor set $\mathcal{R^*}$:
\begin{equation}
    d_{J}(p, g_i) = 1 - \frac{|\mathcal{R}^*(p, k) \cap \mathcal{R}^*(g_i, k)|}{|\mathcal{R}^*(p, k) \cup \mathcal{R}^*(g_i, k)|}.
\end{equation}

The final distance embedding is
\begin{equation}
    d^{*}(p, g_i) = (1-\lambda)d_{J}(p, g_i) + \lambda d^{\prime}(p, g_i).
\end{equation}
\begin{figure}[t]
\begin{center}
   \includegraphics[width=\linewidth]{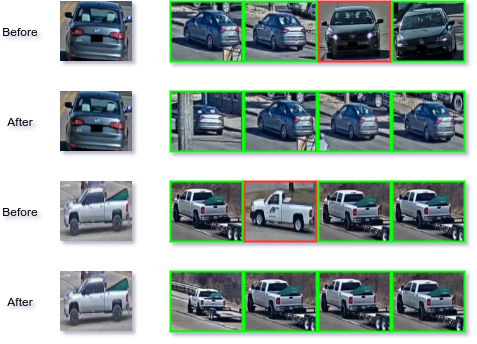}
\end{center}
   \caption{The effects of re-ranking method.}
\label{fig:re-rank}
\end{figure}
\subsubsection{Distance Averaging by Track}
Given the test track for each test image, we calculate the average distance between a probe image $p$ and a track. Then, we replace the distance between the probe image and each image in that track with the calculated average distance. The problem becomes finding tracks that have the most similar car to the probe image instead of finding individual images. The method increases mAP since the top results will be populated with correct images from the same track for uncomplicated cases.
\section{Experiment}
Based on \cite{jakel2019}, we have enough resources for building our models with the provided utilities. After being cropped with Detectron2\cite{wu2019detectron2}, the images are resized to $128 \times 256$ for training GLAMOR models \cite{suprem2020looking} and $224 \times 224$ for training ResNeXt101 model \cite{xie2016aggregated}. Image size may largely affect the re-identification results according to \cite{luo2019bag}; therefore, we choose $128 \times 256$ as our image size because vehicle images tend to have the width larger than the height. The default image size of the pre-trained ResNeXt101 is $224 \times 224$, so we keep it in order to transfer learning efficiently. The images are then augmented with flipping and cropping techniques, color jitter, color augmentation\cite{krizhevsky2017imagenet}, and random erasing \cite{zhong2017random}.

The GLAMOR models are pre-trained with the synthetic data for around $25$ epochs with an initial learning rate of $0.0002$, learning rate decay of $0.2$ for every $10$ epochs, a margin of $0.3$, and the $10:1$ ratio between triplet loss and softmax loss. After that, we feed the transformed images above to the GLAMOR models for the re-identification task with similar parameters. The models converge quickly in around $25$ epochs thanks to the pre-trained weights.

We repeat the same procedure with ResNext101 but with pre-trained weights from ImageNet \cite{krizhevsky2017imagenet, cadene2019}, instead of the synthetic data. After training for $55$ epochs for re-identification task with an initial learning rate of $0.0003$, learning rate decay of $0.3$ for every $20$ epochs, and the same margin and loss ratio, we keep that weight to train the ResNext101 models further to classify type and color. For the classification task, we train the models using softmax loss only with the learning rate of $0.002$ and learning rate decay of $0.5$ for every $20$ epochs.

Even though we have weights of two different models GLAMOR and ResNeXt101 for re-identification tasks, we find that GLAMOR outperforms ResNext101. Therefore, we decide to use only GLAMOR models for the re-identification task. On the other hand, since ResNext101 is a state-of-the-art image classification model, we use it as our metadata attribute extractor.

Table~\ref{table:result} compares the result of our system with those of other teams. Our proposed approach achieves mAP of $37.25$\% and ranks $28$th in Track $2$ of the AI City Challenge 2020. Table~\ref{table:baseline} compares our result with two different base line results provided in \cite{tang2019cityflow}.

\begin{table}
\begin{center}
\begin{tabular}{|c|c|c|c|}
\hline
Rank & Team ID & Team Name & mAP (\%) \\
\hline
$1$ & $73$ & Baidu-UTS & $84.13$ \\
$2$ & $42$ & RuiYanAI & $78.10$ \\
$3$ & $39$ & DMT & $73.22$ \\
$4$ & $36$ & IOSB-VeRi & $68.99$ \\
$5$ & $30$ & BestImage & $66.84$ \\
$6$ & $44$ & BeBetter & $66.83$ \\
... & ... & ... & ... \\
$26$ & $80$ & HCMUS & $38.82$ \\
$27$ & $85$ & MODULABS2 & $38.35$ \\
$\mathbf{28}$ & $\mathbf{4}$ & \textbf{TCU} & $\mathbf{37.25}$ \\
$29$ & $69$ & smiles\_lab & $36.39$ \\
$30$ & $66$ & VPULab@EPS-UAM & $36.23$ \\
$31$ & $59$ & Hsueh-Tse & $35.91$ \\
... & ... & ... & ... \\
$36$ & $100$ & CMU & $24.43$ \\
$37$ & $22$ & psl2020 & $23.68$ \\
$38$ & $50$ & UIT\_NQC & $18.64$ \\
$39$ & $57$ & Insight Centre & $13.00$ \\
$40$ & $62$ & Alavya & $7.59$ \\
$41$ & $75$ & Albany\_NCCU & $3.68$ \\
\hline
\end{tabular}
\end{center}
\caption{Track 2 Competition Results. Our result is highlighted in bold.}
\label{table:result}
\end{table}

\begin{table}
\begin{center}
\begin{tabular}{|c|c|c|}
\hline
Model & Rank@1(\%) & mAP (\%) \\
\hline
ResNet50 & $45.9$ & $29.4$ \\
ResNeXt101 & $48.8$ & $32.0$ \\
\textbf{Ours} & $\mathbf{52.6}$ & $\mathbf{37.3}$ \\
\hline
\end{tabular}
\end{center}
\caption{Comparison with base line models.}
\label{table:baseline}
\end{table}

\section{Conclusion}
In this paper, we introduce an attention-driven re-identification method based on GLAMOR \cite{suprem2020looking}. We also incorporate metadata attribute embedding in the re-ranking process, which boosts the performance of the model. In addition, several techniques in pre-processing and post-processing are adopted to enhance the results. Below are topics that should be further studied in order to improve our system:
\begin{itemize}
    \item Image super-resolution for pre-processing.
    \item GAN-based models in vehicle re-identification.
    \item View-aware feature extraction.
    \item Intensive hyperparameter tuning.
\end{itemize}

{\small
\bibliographystyle{ieee_fullname}
\bibliography{egbib}

\begin{thebibliography}{10}\itemsep=-1pt

\bibitem{cadene2019}
Remi Cadene.
\newblock Pretrained models for {Pytorch}.
\newblock \url{https://github.com/Cadene/pretrained-models.pytorch}, 2019.

\bibitem{chen2019deep}
G. {Chen}, T. {Zhang}, J. {Lu}, and J. {Zhou}.
\newblock Deep meta metric learning.
\newblock In {\em 2019 IEEE/CVF International Conference on Computer Vision
  (ICCV)}, pages 9546--9555, 2019.

\bibitem{deng2009a}
J. {Deng}, W. {Dong}, R. {Socher}, L. {Li}, {Kai Li}, and {Li Fei-Fei}.
\newblock Image{Net}: A large-scale hierarchical image database.
\newblock In {\em 2009 IEEE Conference on Computer Vision and Pattern
  Recognition}, pages 248--255, 2009.

\bibitem{Gou_2017_CVPR_Workshops}
M. Gou, S. Karanam, W. Liu, O. Camps, and R.~J. Radke.
\newblock {DukeMTMC4ReID}: A large-scale multi-camera person re-identification
  dataset.
\newblock In {\em The IEEE Conference on Computer Vision and Pattern
  Recognition (CVPR) Workshops}, July 2017.

\bibitem{he2017maskrcnn}
K. He, G. Gkioxari, P. Doll{\'{a}}r, and R.~B. Girshick.
\newblock Mask {R-CNN}.
\newblock {\em CoRR}, abs/1703.06870, 2017.

\bibitem{hermans2017defense}
A. Hermans, L. Beyer, and B. Leibe.
\newblock In defense of the triplet loss for person re-identification, 2017.

\bibitem{Huang_2019_CVPR_Workshops}
T. Huang, J. Cai, H. Yang, H. Hsu, and J. Hwang.
\newblock Multi-view vehicle re-identification using temporal attention model
  and metadata re-ranking.
\newblock In {\em The IEEE Conference on Computer Vision and Pattern
  Recognition (CVPR) Workshops}, June 2019.

\bibitem{jakel2019}
Jakel21.
\newblock Vehicle {ReID} baseline.
\newblock \url{https://github.com/Jakel21/vehicle-ReID-baseline}, 2019.

\bibitem{Khorramshahi_2019_CVPR_Workshops}
P. Khorramshahi, N. Peri, A. Kumar, A.l Shah, and R. Chellappa.
\newblock Attention driven vehicle re-identification and unsupervised anomaly
  detection for traffic understanding.
\newblock In {\em The IEEE Conference on Computer Vision and Pattern
  Recognition (CVPR) Workshops}, June 2019.

\bibitem{krizhevsky2017imagenet}
A. Krizhevsky, I. Sutskever, and G.~E. Hinton.
\newblock Image{Net} classification with deep convolutional neural networks.
\newblock {\em Commun. ACM}, 60(6):84–90, May 2017.

\bibitem{kumar2019vehicle}
R. Kumar, E. Weill, F. Aghdasi, and P. Sriram.
\newblock Vehicle re-identification: an efficient baseline using triplet
  embedding, 2019.

\bibitem{lin2014microsoft}
T. {Lin}, M. {Maire}, S. {Belongie}, L. {Bourdev}, R. {Girshick}, J. {Hays}, P.
  {Perona}, D. {Ramanan}, C.~L. {Zitnick}, and P. {Dollár}.
\newblock Microsoft {COCO}: Common objects in context, 2014.

\bibitem{liu2016deep}
H. {Liu}, Y. {Tian}, Y. {Wang}, L. {Pang}, and T. {Huang}.
\newblock Deep relative distance learning: Tell the difference between similar
  vehicles.
\newblock In {\em 2016 IEEE Conference on Computer Vision and Pattern
  Recognition (CVPR)}, pages 2167--2175, 2016.

\bibitem{liu2016large-scale}
X. {Liu}, W. {Liu}, H. {Ma}, and H. {Fu}.
\newblock Large-scale vehicle re-identification in urban surveillance videos.
\newblock In {\em 2016 IEEE International Conference on Multimedia and Expo
  (ICME)}, pages 1--6, 2016.

\bibitem{luo2019bag}
H. Luo, Y. Gu, X. Liao, S. Lai, and W. Jiang.
\newblock Bag of tricks and a strong baseline for deep person
  re-identification, 2019.

\bibitem{nguyen2019vehicle}
K. Nguyen, T. Hoang, M. Tran, T. Le, N. Bui, T. Do, V. Vo-Ho, Q. Luong, M.
  Tran, T. Nguyen, T. Truong, V. Nguyen, and M. Do.
\newblock Vehicle re-identification with learned representation and spatial
  verification and abnormality detection with multi-adaptive vehicle detectors
  for traffic video analysis.
\newblock In {\em The IEEE Conference on Computer Vision and Pattern
  Recognition (CVPR) Workshops}, June 2019.

\bibitem{Sankaranarayanan_2016}
S. Sankaranarayanan, A. Alavi, C.~D. Castillo, and R. Chellappa.
\newblock Triplet probabilistic embedding for face verification and clustering.
\newblock {\em 2016 IEEE 8th International Conference on Biometrics Theory,
  Applications and Systems (BTAS)}, Sep 2016.

\bibitem{schroff2015facenet}
F. Schroff, D. Kalenichenko, and J. Philbin.
\newblock Facenet: A unified embedding for face recognition and clustering.
\newblock {\em 2015 IEEE Conference on Computer Vision and Pattern Recognition
  (CVPR)}, Jun 2015.

\bibitem{suprem2020looking}
A. Suprem and C. Pu.
\newblock Looking {GLAMORous}: Vehicle re-id in heterogeneous cameras networks
  with global and local attention, 2020.

\bibitem{szegedy2015rethinking}
C. Szegedy, V. Vanhoucke, S. Ioffe, J. Shlens, and Z. Wojna.
\newblock Rethinking the {Inception} architecture for computer vision, 2015.

\bibitem{tang2019cityflow}
Z. Tang, M. Naphade, M. Liu, X. Yang, S. Birchfield, S. Wang, R. Kumar, D.
  Anastasiu, and J. Hwang.
\newblock {CityFlow}: A city-scale benchmark for multi-target multi-camera
  vehicle tracking and re-identification, 2019.

\bibitem{wang2017residual}
F. Wang, M. Jiang, C. Qian, S. Yang, C. Li, H. Zhang, X. Wang, and X. Tang.
\newblock Residual attention network for image classification, 2017.

\bibitem{wang2017orientation}
Z. {Wang}, L. {Tang}, X. {Liu}, Z. {Yao}, S. {Yi}, J. {Shao}, J. {Yan}, S.
  {Wang}, H. {Li}, and X. {Wang}.
\newblock Orientation invariant feature embedding and spatial temporal
  regularization for vehicle re-identification.
\newblock In {\em 2017 IEEE International Conference on Computer Vision
  (ICCV)}, pages 379--387, 2017.

\bibitem{weinberger2009distance}
K.~Q. Weinberger and L.~K. Saul.
\newblock Distance metric learning for large margin nearest neighbor
  classification.
\newblock {\em J. Mach. Learn. Res.}, 10:207–244, June 2009.

\bibitem{woo2018cbam}
S. Woo, J. Park, J. Lee, and I.~S. Kweon.
\newblock {CBAM}: Convolutional block attention module, 2018.

\bibitem{teng2019}
S. Woo, J. Park, J. Lee, and I.~S. Kweon.
\newblock Official {PyTorch} code for {"BAM: Bottleneck Attention Module
  (BMVC2018)"} and {"CBAM: Convolutional Block Attention Module (ECCV2018)"}.
\newblock \url{https://github.com/Jongchan/attention-module}, 2019.

\bibitem{wu2019detectron2}
Y. Wu, A. Kirillov, F. Massa, W. Lo, and R. Girshick.
\newblock Detectron2.
\newblock \url{https://github.com/facebookresearch/detectron2}, 2019.

\bibitem{xie2016aggregated}
S. Xie, R. Girshick, P. Dollár, Z. Tu, and K. He.
\newblock Aggregated residual transformations for deep neural networks, 2016.

\bibitem{zheng2015scalable}
L. {Zheng}, L. {Shen}, L. {Tian}, S. {Wang}, J. {Wang}, and Q. {Tian}.
\newblock Scalable person re-identification: A benchmark.
\newblock In {\em 2015 IEEE International Conference on Computer Vision
  (ICCV)}, pages 1116--1124, 2015.

\bibitem{zhong2017reranking}
Z. Zhong, L. Zheng, D. Cao, and S. Li.
\newblock Re-ranking person re-identification with k-reciprocal encoding, 2017.

\bibitem{zhong2017random}
Z. Zhong, L. Zheng, G. Kang, S. Li, and Y. Yang.
\newblock Random erasing data augmentation, 2017.

\end{thebibliography}
}

\end{document}